\documentclass[DIV=14, letterpaper]{scrartcl}

\usepackage{amsmath,amssymb,amsfonts,amsthm,subfig}
\usepackage{graphicx}
\usepackage{mathptmx}
\usepackage{cite}
\usepackage{hyperref}

\usepackage{tikz}
\usetikzlibrary{calc}
\usetikzlibrary{matrix}

\def \MM{\mathbb{M}}
\newcommand{\bey}{\begin{eqnarray}}
\newcommand{\eey}{\end{eqnarray}}
\newcommand{\nn}{\nonumber}
\newcommand{\beq}{\begin{equation}}
\newcommand{\eeq}{\end{equation}}
\theoremstyle{plain}

\theoremstyle{definition}

\theoremstyle{remark}

\begin{document}

\title{Anisotropic Mesh Adaptation for Image Representation}
\author{Xianping Li \thanks{
Department of Mathematics and Statistics, the University of Missouri-Kansas City, Kansas City, MO 64110,
U.S.A. ({\tt lixianp@umkc.edu})}
}

\date{}

\maketitle

\begin{abstract}
	
	Triangular meshes have gained much interest in image representation and have been widely used in image processing. This paper introduces a framework of anisotropic mesh adaptation (AMA) methods to image representation and proposes a GPRAMA method that is based on AMA and greedy-point removal (GPR) scheme. Different than many other methods that triangulate sample points to form the mesh, the AMA methods start directly with a triangular mesh and then adapt the mesh based on a user-defined metric tensor to represent the image. The AMA methods have clear mathematical framework and provides flexibility for both image representation and image reconstruction. A mesh patching technique is developed for the implementation of the GPRAMA method, which leads to an improved version of the popular GPRFS-ED method. The GPRAMA method can achieve better quality than the GPRFS-ED method but with lower computational cost.  
	  
\end{abstract}


\section{Introduction}
\label{Sec-intro}

Triangular meshes have recently received considerable interest in adaptive sampling for image representation \cite{DLR90, TV91, DACB96, Ede00, RC01, YMS01, BYG03, YWB03, SW04, BYW04, DDI06, DI06, CGDC07, SD07, BPC09, SD09, Ada11}. One common approach is to find proper sample points then connect the points to form a mesh. For example, Ramponi and Carrato  \cite{RC01} have defined a sample skewness parameter and used a multi-resolution approach to obtain a grid with an almost uniform sample density along the edges and no sample in areas with constant or linearly changing grey level. Yang {\it et al.}  \cite{YWB03} argue that small (in area) elements are needed in image region where the second directional directive is large and have introduced the error diffusion (ED) scheme. They first construct a feature map based on the largest entry (of absolute value) in the Hessian matrix of the image function, then use Floyd-Steinberg dithering scheme to generate sample points, and finally use Delaunay triangulation to connect the nodes into a mesh. Demaret {\it et al.} \cite{DDI06, DI06} have introduced the greedy-point removal (GPR) scheme that first constructs a triangular mesh using all the image points and then removes the sample points that yield smallest reconstruction error repeatedly. Adams \cite{Ada11} has proposed the GPRFS method based on the GPR scheme by replacing the initial triangular mesh of all image points with a subset of the points and developed the GPRFS-ED method that selects the initial points using the ED scheme. 

Another approach is to use a mesh directly to represent the image. For example, Terzopoulos and Vasilescu \cite{TV91} have introduced an adaptive mesh approach where the mesh is considered as a dynamic node/spring system. They sample an image at a reduced rate and then reconstruct it by concentrating the nodes of the mesh at regions where the image values change rapidly (high-gradient region). They develop adaptive meshes with a feedback procedure that automatically adjusts spring parameters according to the observations made at the nodes to which they are attached, and use a Gaussian convolution of the Hessian for the adaptive image reconstruction. Isotropic triangles are used in their adaptive meshes. Courchesne {\it et al.} \cite{CGDC07} use the Hessian matrix based on the gray level of MRI images as a metric tensor to adapt the triangular mesh for 3D reconstruction of human trunk. The Hessian matrix is reconstructed by linear or quadratic fitting. They then adapt the mesh based on the provided metric tensor and four constraint factors - minimum and maximum Euclidean edge lengths, maximum stretching of the metric, and target length of an edge in the metric. Bougleux, Peyre and Cohen \cite{BPC09} have developed a progressive geodesic meshing algorithm that defines a geodesic distance using regularized Hessian as the metric tensor and exploits the anisotropy of images through a farthest point sampling strategy that forces the anisotropic Delaunay triangles to follow the geometry of the image. They have demonstrated the advantages of anisotropic triangular approximation over isotropic triangular approximation. Sarkis and Diepold \cite{SD09} have used binary space partitions in combination with clustering scheme to approximate an image with a mesh. They first cluster the image into a few initial triangles (or rectangles) and then subdivide each triangle (or rectangle) into two or more smaller triangles recursively according to a predefined threshold. 

Most of the adaptive sampling methods are ``content-based'' that use some information from the image such as edges, textures, or Hessian. Different sampling or meshing strategies have been developed. The GPR method provides high quality meshes but requires significant computational cost. On the other hand, the ED method reduces computational cost but provides lower quality meshes. The GPRFS-ED method tries to find a balance between mesh quality and computational cost by combining the advantages of GPR and ED methods. It is worth mentioning that most of the sampling methods take the approach of finding the desired sample points first and then connect the points into a mesh. Only a few methods such as \cite{TV91, CGDC07, BPC09, SD09} follow the approach that starts from an initial mesh and then adapt the mesh to represent the image. 

On the other hand, anisotropic mesh adaptation (AMA) has been successfully applied to improve computational efficiency and accuracy when solving partial differential equations \cite{FA05, HL08, LH10, DZ10, LH13, WDLQV12}. In this paper, we introduce a framework of AMA methods for image representation. AMA methods take the $\MM$-uniform mesh approach for mesh adaptation and use finite element interpolation for image reconstruction. The methods start with an initial triangular mesh, then adapt the mesh according to a user-defined metric tensor $\MM$, and finally reconstruct the image from the mesh. The framework has the flexibility for both mesh adaptation and image reconstruction. Various metric tensors can be chosen for mesh adaptation, and different orders of finite element interpolation can be applied for reconstruction. In this paper, we only consider linear finite element interpolation for triangular elements in the reconstruction step. 

For reader's convenience, the representation methods under consideration are summarized in the following list.
\begin{itemize}
	\item ED: error diffusion method developed by Yang {\it et al.} \cite{YWB03}. 
	\item AMA methods such as $\MM_{aniso, k}$: anisotropic mesh adaptation method using metric tensor $\MM_{aniso}$ with the initial mesh being adapted $k$ times to generate the desired mesh, proposed in Section \ref{Sec-AMArep}.	
	\item GPR: greedy-point removal scheme proposed by Demaret {\it et al.} \cite{DDI06, DI06}.
	\item GPRFS-ED: modified GPR scheme starting from a subset of points chosen by ED method, proposed by Adams \cite{Ada11}.
	\item GPRED-CDT($\gamma$): GPR starting from $\gamma$ times of the desired number of sample points chosen by ED and utilizing constrained Delaunay triangulation for mesh patching, essentially the same as GPRFS-ED, proposed in Section \ref{Sec-GPRAMA}.
	\item GPRED-EC($\gamma$): same as GPRED-CDT($\gamma$) except using Ear Clipping for mesh patching, proposed in Section \ref{Sec-GPRAMA}.
	\item GPRAMA($\gamma$): GPR starting from an AMA representation of $\gamma$ times of the desired sample density and utilizing Ear Clipping for mesh patching, proposed in Section \ref{Sec-GPRAMA}.
\end{itemize}

The remainder of this paper is organized as follows. Firstly, a brief introduction of the AMA methods is given in Section \ref{Sec-AMAintro} where the details of the methods can be found in \cite{Hua05b, Hua06, HR11}. Then in Section \ref{Sec-AMArep}, the AMA representation framework is introduced and some results obtained from different methods are presented. In Section \ref{Sec-GPRAMA}, a GPRAMA representation method based on AMA and GPR is proposed and some results and computational complexity are discussed. Finally, some conclusions and comments are given in Section \ref{Sec-con}. For reader's convenience, a brief summary of finite element interpolation for triangular elements is provided in the Appendix.

\section{Anisotropic mesh adaptation (AMA) methods}
\label{Sec-AMAintro}

Different adaptive sampling methods and mesh strategies have been applied in image representation by other researchers as summarized in Section \ref{Sec-intro}. In this section, we introduce the ``anisotropic mesh adaptation'' (AMA) methods. AMA methods take the $\MM$-uniform mesh approach, with which an adaptive mesh is generated as a uniform mesh in the metric specified by a tensor $\MM$. The metric tensor $\MM$ is required to be strictly positive definite and it determines the size, shape and orientation of the triangular elements \cite{Hua06}. Once a metric tensor is specified, the free C++ code BAMG (Bidimensional Anisotropic Mesh Generator) developed by Hecht \cite{bamg} is used to generate the corresponding triangular mesh.  
BAMG first generates an initial mesh based on the geometry of the domain provided in a file that defines the nodes and edges and the desired mesh size using constrained Delaunay triangulation. Then users have the choice to either provide a metric tensor on the initial background mesh or use the internal metric tensor computed by BAMG. Once a metric tensor $\MM$ is provided, BAMG employs five local minimization tools including edge suppression, vertex suppression, vertex addition, edge swapping, and vertex relocation to generate the desired anisotropic mesh according to $\MM$.
One of our objectives in this paper is to build the framework for AMA in image representation that can take different metric tensors for different needs in image processing. 

Firstly, we introduce some notations and the conditions for $\MM$-uniform meshes.
Let $\Omega$ be the spatial domain, $K$ be any triangular element in a simplicial mesh $\mathcal{T}_h$, and $\hat{K}$ to be the reference element that is equilateral and unitary in area. Let $F_K$ be the affine mapping from $\hat{K}$ to $K$. 
An $\MM$-uniform 2-D triangular mesh $\mathcal{T}_h$ for a given metric tensor $\MM=\MM({\bf x})$ satisfies the following condition
\beq
\label{cond-both}
(F_K')^T \MM_K F_K' = \frac{\sigma_h}{N} I, \quad \forall K \in \mathcal{T}_h 
\eeq
that is equivalent to the following two conditions \cite{Hua06}
\bey
\label{cond-eq}
|K| \sqrt{\mbox{det}(\MM_K)} & = & \frac{\sigma_h}{N}, \quad \forall K \in \mathcal{T}_h, \\
\label{cond-ali}
\frac{1}{2} \mbox{tr} \left ( (F_K')^T \MM_K F_K' \right ) & = &
\mbox{det} \left ( (F_K')^T \MM_K F_K' \right )^{\frac{1}{2}}, \quad \forall K \in \mathcal{T}_h,
\eey
where $I$ is the identity matrix of size $2 \times 2$, $|K|$ is the area of the element $K$, $N$ is the number of mesh elements,  
$F_K'$ is the Jacobian matrix of $F_K$,
\beq
\MM_K = \frac{1}{|K|} \int_K \MM({\bf x}) d {\bf x}, \mbox{ and }
\sigma_h = \sum_{K \in \mathcal{T}_h} |K| \sqrt{\mbox{det}(\MM_K)}.
\eeq
Condition (\ref{cond-eq}) is called the {\em equidistribution condition} and determines the size of element $K$, while condition (\ref{cond-ali}) is called the {\em alignment condition} and characterizes the shape and orientation of $K$.

In the framework of AMA methods, the goal is to develop and use proper metric tensors based on the needs of the problems. Different metric tensors will have different properties and features. It is worth mentioning that Hessian matrix $H$ is not an optimal metric tensor \cite{HR11} and may not be positive definite. In our framework, we replace the Hessian with its absolute form defined as follows
\beq
\label{H-abs}
|H|= Q \left[\begin{array}{cc} |\lambda_1| & 0 \\ 0 & |\lambda_2| \end{array} \right] Q^{-1}, \text{ with } H= Q \left[\begin{array}{cc} \lambda_1 & 0 \\ 0 & \lambda_2 \end{array} \right] Q^{-1},
\eeq
where $\lambda_1$ and $\lambda_2$ are the eigvenvalues of $H$ and $Q$ is the matrix of the corresponding eigenvectors. 
The metric tensor $|H|$ is denoted as $\MM_{H}$ in this paper, and \cite{CGDC07} can be considered as a specific example in our AMA framework. Some other metric tensors are described below.

For isotropic mesh adaptation, a metric tensor $\MM_{iso}$ is defined for any triangular element $K$ as follows \cite{Hua05b}
\beq
\label{M-iso}
\MM_{iso,K}=\left( 1 + \frac{1}{\alpha_h}  \| H_K \|_F \right) \, I ,
\eeq
where $H_K$ denotes the value of $H$ at the center of element $K$, $\|\cdot\|_F$ is the Frobenius matrix norm, and $\alpha_h$ is a regularization factor that is defined by
\beq
\alpha_h = \frac{1}{| \Omega |} \left(  \sum_{K\in \mathcal{T}_h} |K| \cdot  \| H_K \|_F \right).
\eeq
$\MM_{iso}$ provides isotropic mesh adaptation where all triangles are of the same shape but may have different sizes, and more triangles will be concentrated in the high-gradient region.

For anisotropic mesh adaptation, a metric tensor $\MM_{aniso}$ is developed in \cite{Hua05b} that is based on minimization of a bound on the $H^1$ semi-norm of linear interpolation error and is defined for any triangular element $K$ as follows
\beq
\label{M-aniso}
\MM_{aniso,K}=\rho_K \det
\left( I+\frac{1}{\alpha _{h}}|H_K|\right) ^{-\frac{1}{2}}
\left[ I+\frac{1}{\alpha _{h}}|H_K|\right] ,
\eeq
and 
\beq
\rho_K = \Big \| I + \frac{1}{\alpha_h} | H_K | \Big \|_F ^{\frac{1}{2}
}\,\det \left( I+\frac{1}{\alpha _{h}}|H_K|\right)^{\frac{1}{4}},
\eeq
where $\alpha_h$ is the regularization parameter and is defined implicitly through
\beq
\sum_{K\in \mathcal{T}_h} \rho_K |K| = 2 | \Omega | .
\eeq
With this choice of $\alpha_h$, roughly fifty percents of the triangular elements will be concentrated in large gradient regions \cite{Hua05b}. The adaptation is anisotropic because the triangles in the mesh may have different size, shape and orientation. 

For image processing with anisotropic diffusion filters \cite{Wei98}, a metric tensor $\MM_{DMP}$ is developed in \cite{LH10} that takes the inverse of the diffusion tensor. The elements of the mesh based on $\MM_{DMP}$ will be aligned along the principle diffusion direction, and the corresponding numerical solution will satisfy the maximum principle under some conditions of time step \cite{LH13}. Another metric tensor $\MM_{DMP+adap}$ is also developed in \cite{LH10} that combines the properties of both $\MM_{aniso}$ and $\MM_{DMP}$, that is, the mesh not only provides numerical solution that satisfies maximum principle but also performs adaptation based on the interpolation error.

In this paper, we only focus on the metric tensors $\MM_{H}$, $\MM_{iso}$, and $\MM_{aniso}$. In the computations for those metric tensors, the Hessian matrix $H$ at a point is reconstructed by the least-squares fitting from function values at neighboring vertices. For convenience, we use the metric tensor to denote the mesh as well as the corresponding representation. For example, $\MM_{aniso}$ denotes the mesh and representation according to the metric tensor $\MM_{aniso}$. 

As demonstrated in \cite{BPC09, HR11}, anisotropic meshes have advantages over isotropic meshes in terms of computational efficiency and accuracy. Our results in Section \ref{Sec-AMArep} also confirm that anisotropic meshes provide better representation quality than isotropic meshes. Therefore, we will only use anisotropic mesh adaptation methods in the AMA image representation framework.

\section{AMA image representation framework}
\label{Sec-AMArep}

In this section, we introduce the AMA framework for image representation. We consider an image as a function $f$ that is defined on a set $\Lambda$ of points on domain $\Omega=[0,1]\times[0,1]$. Let $S$ denote the set of desired sample points and $SD$ denote the sample density that is defined as 
\beq
SD=|S|/|\Lambda|,
\eeq
where $|\cdot|$ is the cardinality of the set. The quality of the mesh (or representation) is measured by the peak-signal-to-noise-ratio (PSNR) that is calculated in decibels (dB) as follows \cite{Ada11} 
\beq
\text{PSNR} = 20 \log_{10} \left( \frac{2^p-1}{d} \right), \quad 
d = \left( \frac{1}{|\Lambda|} \sum_{i \in \Lambda} | \hat{f}(i)-f(i) |^2 \right)^{\frac{1}{2}},
\eeq
where $\hat{f}$ is the reconstructed image from the triangular mesh, and $p$ is the sample precision in bits/sample. Larger value of PSNR indicates better mesh quality (or representation).

\subsection{Framework}
\label{Subsec-AMArep}

In this paper, we apply the AMA methods for image representation, in which a triangular mesh with fewer points is used to represent the original image $f$. Given the original image $f$, we use the free C++ code BAMG \cite{bamg} to generate an initial triangular mesh with number of vertices $N_v$ that is much smaller than $|\Lambda|$. Then the values on the vertices are interpolated from $f$, and the metric tensor $\MM$ is computed for each triangle in the initial mesh. With the computed metric tensor, BAMG generates the desired anisotropic mesh using the procedures described in Section \ref{Sec-AMAintro}. Finally, the image is reconstructed from the final mesh using finite element interpolation.

It is difficult to generate a mesh that satisfies conditions \eqref{cond-eq} and \eqref{cond-ali} exactly. Moreover, the initial mesh, in general, may not contain the important information from the original image. Therefore, the mesh can be adapted multiple times in order to obtain a final mesh that is close to be an $\MM$-uniform mesh, or the so-called quasi-$\MM$-uniform mesh. The iteration can be terminated if further adaptation does not significantly improve image quality (measured by PSNR), and can be image dependent in order to obtain the best representation of a particular image. However, numerical results show that 2 to 5 iterations are sufficient to provide a quasi-$\MM$-uniform mesh with good quality, and further adaptation does not improve the quality significantly.  

More specifically, the AMA representation framework consists of the following four steps.

{\bf Step 1}: Generate an initial mesh based on the desired sample density.

{\bf Step 2}: Assign function values to mesh vertices (and interpolation nodes) from original image using linear finite element interpolation and compute the user-defined metric tensor $\MM$ on the mesh.  

{\bf Step 3}: Adapt the mesh to be a quasi-$\MM$-uniform mesh that almost fits the provided metric tensor $\MM$. 

{\bf Step 4}: Reconstruct the image using the final quasi-$\MM$-uniform mesh with finite element interpolation for triangles. 

\begin{figure}[ht!]
\centering
\includegraphics[width=6in]{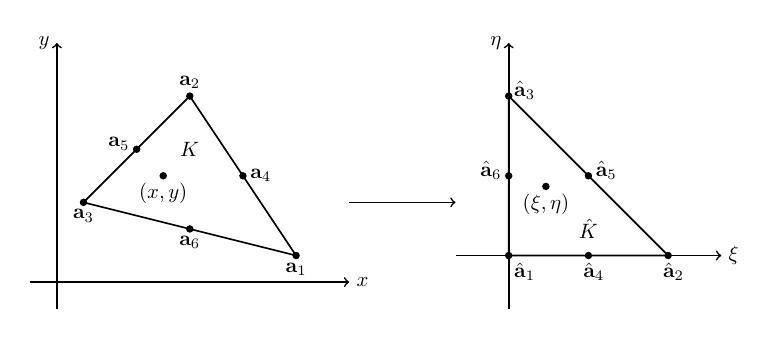}
\caption{Sketch of triangular element $K$ and its reference element $\hat K$ for finite element interpolation, where $\hat K$ is an isosceles right triangle with vertices $\hat{{\bf a}}_1 (0,0)$, $\hat{{\bf a}}_2 (1,0)$, and $\hat{{\bf a}}_3 (0,1)$.}
\label{feminter}
\end{figure}

During the reconstruction step (Step 4), for a particular image pixel, we first locate the triangle that the pixel lies on or in. Then we compute the coordinates of the pixel in the reference element (see Fig. \ref{feminter}) and the corresponding basis functions at the interpolation nodes. Finally, we interpolate the function value from the interpolation nodes using the basis functions as the weights. For linear interpolation, only the three vertices are needed, while for quadratic interpolation, the midpoints are also needed. For reader's convenience, a brief summary of finite element interpolation for triangular elements is provided in the Appendix.

\begin{figure}[ht!]
\centering
\includegraphics[width=6in]{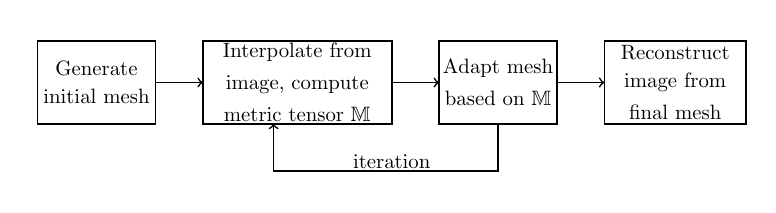}
\caption{Procedures for AMA representation method based on metric tensor $\MM$.}
\label{steps}
\end{figure}

The above procedures are shown in Fig. \ref{steps} where Step 2 and Step 3 can be repeated multiple times in order to obtain better results. For convenience, we denote the number of iterations for Step 2 and Step 3 by $k$, and the corresponding mesh as $\MM_{k}$. For example, for metric tensor $\MM_{H}$, the representation is denoted as $\MM_{H,k}$ if there are $k$ iterations of Step 2 and Step 3. For metric tensor $\MM_{aniso}$, the corresponding mesh is denoted as $\MM_{aniso,k}$. When $k=1$, the mesh is only adapted from initial mesh once and no further adaptation is performed. As mentioned before, we take $k \in [2,5]$ in our computations. 

Note that we can start with a random initial mesh that has more number of vertices than desired ($N_v > |S|$) in Step 1, then iterate Step 2 and Step 3 to obtain a mesh with desired sample density. By this way, more information from the original image could be reserved by the mesh. 
Another approach is to obtain an initial mesh with $N_v > |S|$ for Steps 1 to 3, then reduce the number of mesh vertices to the desired number $|S|$ using the GPR algorithm before Step 4. The particular representation using GPR before Step 4 is denoted as GPRAMA and will be discussed in Section \ref{Sec-GPRAMA}.  

Note that in Step 4, we can choose different orders of finite element interpolation methods. However, in this paper, we only consider linear finite element interpolation for triangular elements and the effects of higher order interpolation on representation quality is currently under investigation. In fact, quadratical interpolation provides higher representation quality, however, the sample density is also higher since it uses the midpoints on the edges of the triangles. There is no need to sample the midpoints because their coordinates can be computed, however, the function values at the midpoints need to be assigned in Step 2. A fair comparison is needed between higher order interpolation and linear interpolation with the same sample density.

\subsection{Results}

\begin{figure}[ht!]
\centering
\includegraphics[width=6in]{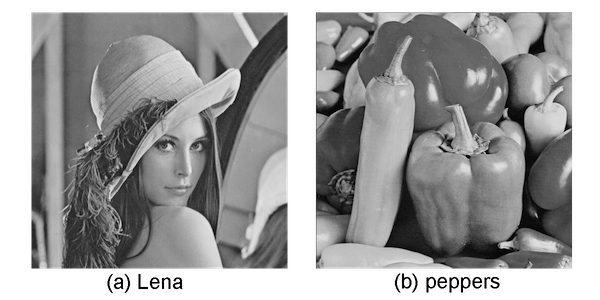}
\caption{Images from USC-SIPI Image Database \cite{USC-SIPI} with pixel resolution $512 \times 512$: (a) Lena; (b) peppers. The RGB components of each pixel are converted to luminance using the weighted sum $0.2989 \cdot R+0.5870 \cdot G+0.1140 \cdot B$.}
\label{img-ini}
\end{figure}

For evaluation purpose, we take the two widely used images, ``Lena'' and ``peppers'', available from USC-SIPI Image Database \cite{USC-SIPI}. 
Fig. \ref{img-ini} shows the initial images of Lena and peppers with pixel resolution $512 \times 512$, while the RGB components of each pixel are converted to greyscale luminance using the weighted sum $0.2989 \cdot R+0.5870 \cdot G+0.1140 \cdot B$. 
Three more images with different resolutions and features are also tested for comparison purpose, including ``roof'', ``lighthouse'' and ``saturn''. Image roof has resolution $1024 \times 1024$ and is obtained from USC-SIPI Image Database \cite{USC-SIPI}. ``lighthouse'' has resolution $480 \times 640$ and ``saturn'' has resolution $1500 \times 1200$, both are taken from MATLAB R2016a imagedata folder. All images are converted to greyscale images as done for images Lena and peppers.

\begin{figure}[ht!]
\centering
\includegraphics[width=6in]{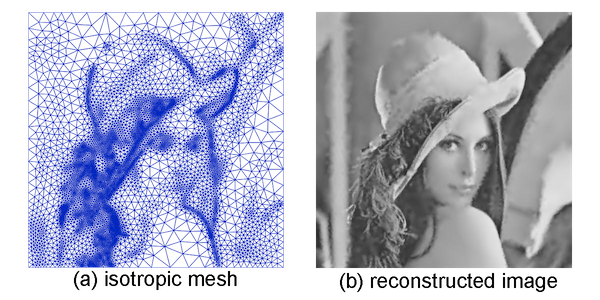}
\caption{Representation of image Lena at sample density of 3\% using $\MM_{iso,3}$: (a) triangular mesh $\MM_{iso,3}$; (b) reconstructed image, PSNR=28.47.}
\label{img-iso-lena}
\end{figure}

\begin{figure}[ht!]
\centering
\includegraphics[width=6in]{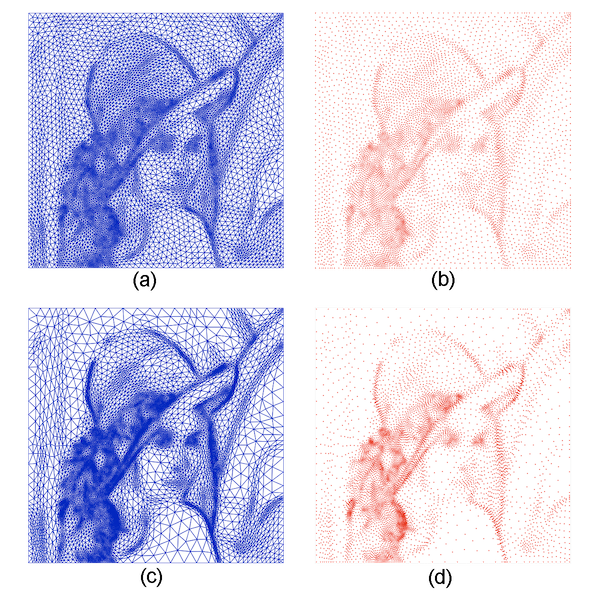}
\caption{Meshes and sample points of image Lena at sample density of 3\% using $\MM_{aniso}$: (a) $\MM_{aniso,1}$ mesh; (b) sample points from the mesh in (a); (c) $\MM_{aniso,2}$ mesh; (d) sample points from the mesh in (c).}
\label{img-aniso-lena}
\end{figure}

\begin{figure}[ht!]
\centering
\includegraphics[width=6in]{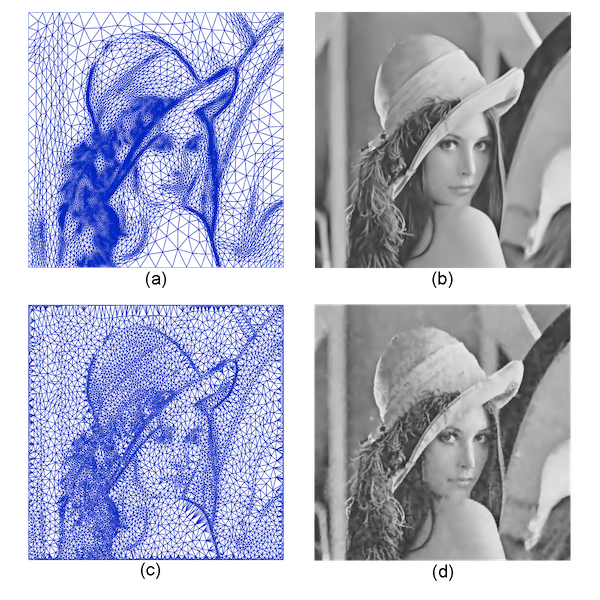}
\caption{Representations of image Lena at sample density of 3\% using different methods: (a) triangular mesh $\MM_{aniso,3}$; (b) reconstructed image from (a), PSNR=31.00; (c) mesh obtained using ED scheme; (d) reconstructed image from (c), PSNR=28.41.}
\label{img-rep-lena}
\end{figure}

Fig. \ref{img-iso-lena} shows the representation of the image Lena at $SD=3\%$ using isotropic mesh according to $\MM_{iso,3}$, and the quality of representation is PSNR=28.26. Fig. \ref{img-aniso-lena} shows the $\MM_{aniso,1}$ and $\MM_{aniso,2}$ meshes and the corresponding sample points of the image Lena at $SD=3\%$. The representation quality for $\MM_{aniso,1}$ is PSNR=29.81, and is PSNR=30.81 for $\MM_{aniso,2}$. After three iterations, the quality increases to PSNR=31.00 for $\MM_{aniso,3}$ as shown in Fig. \ref{img-rep-lena}. Further adaptation does not improve the representation quality for this case. It is clear that image representation based on $\MM_{aniso}$ is better than the one based on $\MM_{iso}$. 

Fig. \ref{img-rep-lena} also shows the representation and reconstruction of the image Lena at $SD=3\%$ using ED scheme denoted as ED. For the ED scheme, we have applied the strategies recommended in \cite{Ada11}, including $B(3)$ smoothing for image data, zero extension for boundary points, and serpentine scan order for the error diffusion. The representation quality is PSNR=28.41 for ED. 

\begin{figure}[ht!]
\centering
\includegraphics[width=6in]{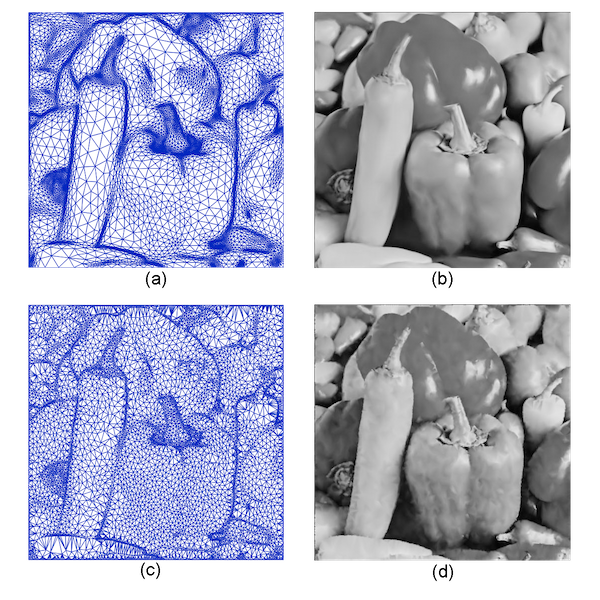}
\caption{Representations of image peppers at sample density of 3\% using different methods: (a) triangular mesh $\MM_{aniso,3}$; (b) reconstructed image from (a), PSNR=30.89; (c) mesh obtained using ED scheme; (d) reconstructed image from (c), PSNR=28.05.}
\label{img-rep-pepper}
\end{figure}

Fig. \ref{img-rep-pepper} shows similar results for the image peppers at $SD=3\%$. The quality for $\MM_{aniso,3}$ is PSNR=30.89 and is PSNR=28.05 for ED. As can be seen from Fig. \ref{img-rep-lena} and Fig. \ref{img-rep-pepper}, the $\MM_{aniso,3}$ meshes preserve key features of the original image by concentrating more triangular elements around the edges and texture regions. The quality of $\MM_{aniso,3}$ representation is much better than that of the ED representation.

The representation qualities for both image Lena and peppers using different AMA methods are shown in Table \ref{table-AMA}. The results obtained using ED scheme are also presented for comparison purpose. As can be seen, the quality of $\MM_{iso,3}$ representation is comparable to ED but not as good as the anisotropic ones. 

\begin{table}[thb]
\caption{Comparison of mesh qualities obtained with various methods}
\vspace{2pt}
\centering
\begin{tabular}{|c|p{10mm}||c|c|c|c|c|c|c|}
\hline
& Sample & \multicolumn{6}{c|}{PSNR (dB)} \\ \cline{3-8} 
Image & Density (\%) & ED & $\MM_{iso,3}$ & $\MM_{H,1}$ & $\MM_{aniso,1}$ & $\MM_{H,3}$ & $\MM_{aniso,3}$ \\
\hline
 & 1.0 & 21.38 & 24.13 & 25.42 & 25.57 & 26.32 & {\bf 26.51} \\ 
 & 2.0 & 26.36 & 26.55 & 28.08 & 28.38 & 28.93 & {\bf 29.35} \\ 
Lena & 3.0 & 28.41 & 28.26 & 29.67 & 29.81 & 30.45 & {\bf 31.00} \\ 
 & 4.0 & 29.77 & 29.42 & 30.71 & 30.83 & 31.82 & {\bf 31.99} \\ 
 & 6.0 & 31.46 & 31.06 & 32.09 & 32.26 & 33.22 & {\bf 33.31}  \\ 
\hline
 & 1.0 & 20.66 & 23.50 & 25.01 & 25.69 & {\bf 26.31} & 25.80 \\ 
 & 2.0 & 25.30 & 26.10 & 28.16 & 28.55 & 29.34 & {\bf 29.36} \\ 
peppers & 3.0 & 28.05 & 28.31 & 29.65 & 30.01 & {\bf 31.05} & 30.89 \\ 
 & 4.0 & 29.57 & 29.14 & 30.74 & 30.94 & {\bf 31.86} & {\bf 31.86} \\ 
 & 6.0 & 31.03 & 31.11 & 31.90 & 32.01 & {\bf 33.10} & 32.95  \\ 
\hline 
\end{tabular} \\
\vspace{2pt}
\label{table-AMA}
\end{table}

The results confirm that more iterations of Step 2 and Step 3, that is, increasing the values of $k$, does improve the mesh quality. For example, for image Lena at $SD=3\%$, PSNR increases from 29.67 for $\MM_{H,1}$ to 30.45 for $\MM_{H,3}$ and from 29.81 for $\MM_{aniso,1}$ to 31.00 for $\MM_{aniso,3}$. Similar results are observed for image peppers. The reason is that after each mesh adaptation, better information are preserved from the original image and the mesh is closer to the $\MM$-uniform mesh. However, for the two images we are investigating, $k=3$ already provides a mesh with good quality and further adaptation does not make significant improvement. The results of PSNR values using $\MM_{H,k}$ and $\MM_{aniso,k}$ at different $k$ values are presented in Table \ref{table-kvalue} for different images at sample density of 3\%. The optimal value of $k$ depends on the given image. For example, $k=2$ is the best for image saturn, $k=3$ works the best for images Lena, and peppers. For image lighthouse, $\MM_{H,k}$ and $\MM_{aniso,k}$ have different optimal $k$ values and $k=3$ is a good balance.

\begin{table}[thb]
\caption{PSNR (dB) of $\MM_{H,k}$ and $\MM_{aniso,k}$ representations at sample density of 3\%}
\vspace{2pt}
\centering
\begin{tabular}{|c|c||c|c|c|c|c|}
\hline
Image & Mesh & \multicolumn{5}{c|}{PSNR (dB)} \\ \cline{3-7} 
 &  & $k=1$ & $k=2$ & $k=3$ & $k=4$ & $k=5$ \\
\hline
Lena & $\MM_{H,k}$ & 29.67 & 29.97 & {\bf 30.45} & 30.19 & 30.18 \\ 
 & $\MM_{aniso,k}$ & 29.81 & 30.48 & {\bf 31.00} & 30.01 & 29.80 \\ 
\hline
lighthouse & $\MM_{H,k}$ & 26.54 & 26.68 & 26.76 & {\bf 26.85} & 26.68 \\ 
 & $\MM_{aniso,k}$ & 26.69 & {\bf 26.85} & 26.78 & 26.62 & 26.19 \\ 
\hline
peppers & $\MM_{H,k}$ & 29.65 & 30.16 & {\bf 31.05} & 29.93 & 29.48 \\ 
 & $\MM_{aniso,k}$ & 30.01 & 29.74 & {\bf 30.89} & 29.92 & 29.69 \\ 
\hline
roof & $\MM_{H,k}$ & 27.40 & 28.07 & 28.43 & 28.66 & {\bf 28.79} \\ 
 & $\MM_{aniso,k}$ & 27.55 & 28.16 & 28.48 & 28.62 & {\bf 28.79} \\ 
\hline
saturn & $\MM_{H,k}$ & 48.89 & {\bf 49.93} & 48.74 & 46.05 & 45.91 \\ 
 & $\MM_{aniso,k}$ & 48.97 & {\bf 49.89} & 49.77 & 48.87 & 47.91 \\ 
\hline
\end{tabular} \\
\vspace{2pt}
\label{table-kvalue}
\end{table}

Comparing $\MM_{H,1}$ and $\MM_{aniso,1}$ in Table \ref{table-AMA} and Table \ref{table-kvalue}, we see that the absolute Hessian $|H|$ is not an optimal metric tensor, and the performance of $\MM_{aniso,1}$ is better than $\MM_{H,1}$ for all cases, although the difference is not significant. By adapting both $\MM_{H}$ and $\MM_{aniso}$ meshes three times, $\MM_{aniso,3}$ performs better than $\MM_{H,3}$ for image Lena while the opposite occurs for image peppers. The results of $\MM_{H,3}$ and $\MM_{aniso,3}$ for the other three images are presented in Table \ref{table-AMA2}. 
The qualities of the representation depend on the specific image but overall performance are comparable for $\MM_{H}$ and $\MM_{aniso}$. In this paper, we choose $\MM_{aniso,3}$ as the representative from the AMA framework and propose a new image representation method based on AMA and GPR in the next section.

\begin{table}[thb]
\caption{Comparison of mesh qualities obtained with $\MM_{H,3}$ and $\MM_{aniso,3}$}
\vspace{2pt}
\centering
\begin{tabular}{|c|p{20mm}||c|c|c|}
\hline
& Sample & \multicolumn{3}{c|}{PSNR (dB)} \\ \cline{3-5} 
Image & Density (\%) & ED & $\MM_{H,3}$ & $\MM_{aniso,3}$ \\
\hline
 & 1.0 & 20.16 & 23.49 & {\bf 23.51} \\ 
 & 2.0 & 24.10 & 25.40 & {\bf 25.58} \\ 
lighthouse & 3.0 & 25.69 & 26.76 & {\bf 26.78} \\ 
 & 4.0 & 26.79 & 27.69 & {\bf 27.84} \\
 & 6.0 & 28.49 & 29.15 & {\bf 29.22} \\ 
\hline
 & 1.0 & 19.19 & 24.90 & {\bf 25.31} \\ 
 & 2.0 & 23.39 & 27.07 & {\bf 27.42} \\ 
roof & 3.0 & 26.61 & 28.43 & {\bf 28.48} \\ 
 & 4.0 & 28.02 & {\bf 29.33} & {\bf 29.33} \\ 
 & 6.0 & 29.24 & {\bf 30.41} & 30.38 \\
\hline
 & 1.0 & 42.14 & 47.72 & {\bf 47.99} \\ 
 & 2.0 & 46.18 & 49.07 & {\bf 49.32} \\ 
saturn & 3.0 & 47.33 & 48.74 & {\bf 49.77} \\ 
 & 4.0 & 48.18 & 48.94 & {\bf 49.74} \\
 & 6.0 & 48.48 & {\bf 47.92} & 47.74 \\ 
\hline 
\end{tabular} \\
\vspace{2pt}
\label{table-AMA2}
\end{table}

\section{GPRAMA representation method}
\label{Sec-GPRAMA}

In this section, we apply the greedy-point removal scheme to AMA representation and propose a new method denoted as GPRAMA. Adams has proposed the GPRFS method in \cite{Ada11} that is based on the GPR scheme while replacing the initial triangular mesh of all image points with a subset $S_0 \subseteq \Lambda$. The GPRFS method starts with $|S_0|= \gamma |S|$ for $\gamma \in [4,5.5]$ and then uses GPR scheme to reduce the number of points from $|S_0|$ to the desired number $|S|$. Adams employs ED method to choose $S_0$ and denotes the method as GPRFS-ED in his paper.  

\begin{figure}[ht!]
\centering
\includegraphics[width=6in]{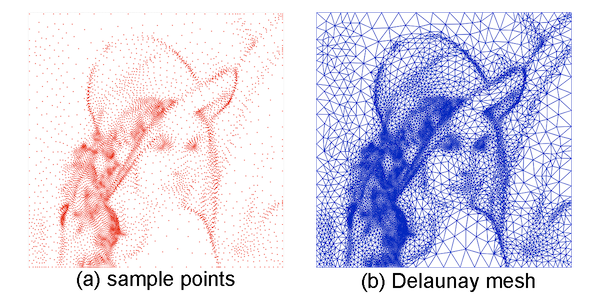}
\caption{Representation of image Lena at sample density of 3\%: (a) sample points; (b) Delaunay mesh, PSNR=28.57.}
\label{img-AMA-Del}
\end{figure}

As discussed in the previous section, our AMA representation methods provide better quality than ED scheme. Therefore, it is reasonable to consider the vertices of an AMA mesh as the initial subset $S_0$ for the GPR scheme. In this sense, the GPRAMA method is a specific example of the GPRFS method. However, GPRFS method relies on Delaunay triangulation of the sample points that does not have the anisotropic feature as in AMA meshes. In fact, the mesh quality obtained via Delaunay triangulation may not be optimal for a given set of sample points, measured by PSNR value of the reconstructed image. Figure \ref{img-AMA-Del} shows one example, where the mesh is obtained by Delaunay triangulation of the provided sample points and the mesh quality is PSNR=28.57; however, for the same set of sample points, the $\MM_{aniso,3}$ mesh (see Fig. \ref{img-rep-lena}(a)) has better quality with PSNR=31.00. 

\subsection{Mesh patching technique}
\label{Subsec-meshpat}

\begin{figure}[ht!]
\centering
\includegraphics[width=6in]{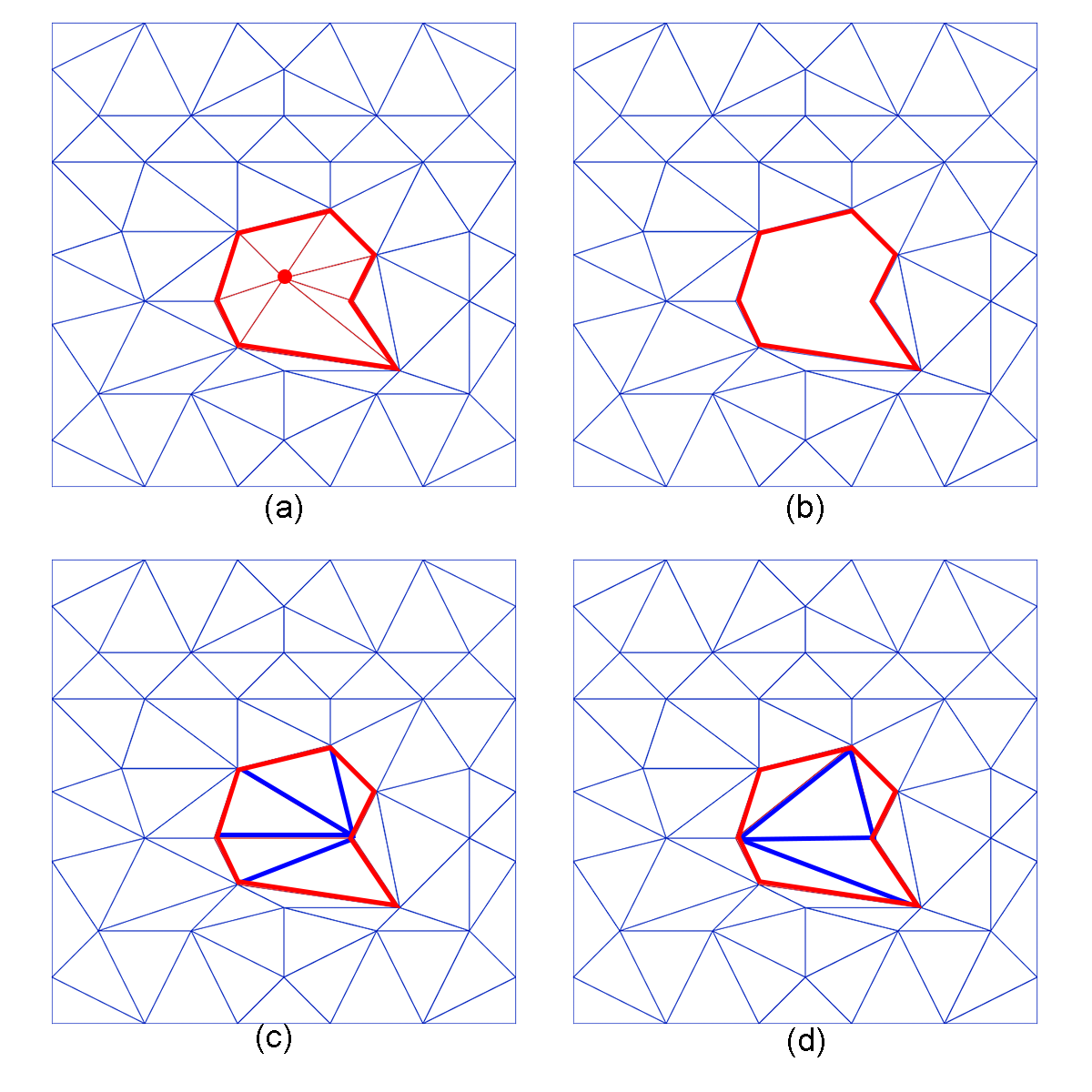}
\caption{Illustration of mesh patching technique: (a) original mesh with the highlighted patch of the point to be removed; (b) mesh with polygon of the empty patch; (c) mesh patching using constrained Delaunay triangulation (CDT) for (b); (d) mesh patching using Ear Clipping (EC) method for (b).}
\label{img-patching}
\end{figure}

In order to preserve the anisotropic feature of the AMA meshes when applying GPR scheme, we have developed a mesh patching technique that attaches a local mesh to the AMA mesh. This patching technique also works for any other triangular meshes. Let $i$ be the index of a general vertex in the mesh, we denote the region covered by triangles sharing vertex $i$ including the boundary edges by $\overline{\omega_i}$ and call it the patch of $i$. The polygon surrounding $i$ formed by the boundary edges, is denoted as $\partial \omega_i$, and the inner region of the patch, that is, excluding the boundary edges, is denoted as $\omega_i$. Before removing the vertex $i$, the patch $\overline{\omega_i}$ is partitioned by the triangular elements from the initial mesh, and linear finite element interpolation are used on those triangles for image reconstruction. If vertex $i$ is chosen to be removed, $\overline{\omega_i}$ is triangulated again without vertex $i$. The new triangulation of $\overline{\omega_i}$ is then added to the global mesh structure. Figure \ref{img-patching} provides an illustration of the mesh patching technique with two different triangulation methods of the patch - one is the constrained Delaunay triangulation (CDT) \cite{JS02} and the other is the Ear Clipping (EC) method \cite{JR98}. Both CDT and EC work well for general polygons including concave ones. The specific procedures for the mesh patching of $\overline{\omega_i}$ are as follows. 

{\bf Step 1}: Save the global indices of the vertices of the polygon $\partial \omega_i$.

{\bf Step 2}: Triangulate $\overline{\omega_i}$ without vertex $i$ (using CDT or EC) and save the mesh connectivity that lists the local indices of the vertices of each triangle. 

{\bf Step 3}: Map the local indices in Step 2 to the global indices from Step 1, and add the updated connectivity information of $\overline{\omega_i}$ to the global mesh.

\subsection{GPRAMA method}
\label{Subsec-GPRAMA}

With the mesh patching technique described above and the AMA representation framework introduced in Section \ref{Sec-AMArep}, the GPRAMA method consists of the following four steps. 

{\bf Step 1}: Generate an AMA representation based on a metric tensor $\MM$ with number of vertices $N_v= \gamma |S|$ for $\gamma \ge 1$. Let $V$ be the set of all vertices in the mesh and let $V_p = V$.

{\bf Step 2}: For any vertex $v_i \in V_p$, compute the significance measure $\delta e_{i}$ defined as the difference between the local mean square error of $\overline{\omega_i}$ after and before removing $v_i$, as shown below.
\beq
\delta e_{i} =  \sum_{j \in \Lambda \cap \overline{\omega_i} } | \hat{f}_a (j)-f(j) |^2 - 
\sum_{j \in \Lambda \cap \overline{\omega_i} } | \hat{f}_b (j)-f(j) |^2,
\eeq 
where $\hat{f}_a$ is the reconstructed value using the new triangulation of the patch without vertex $i$ and $\hat{f}_b$ is the one with vertex $i$. 

{\bf Step 3}: For the vertex $v_i \in V$ having minimal $\delta e_{i}$, reset $V_p$ as the set of vertices of the polygon $\partial \omega_i$, remove $v_i$ and apply the mesh patching technique for $\overline{\omega_i}$. Reset $V=V \backslash \{v_i\}$ and $N_v = N_v - 1$. 

{\bf Step 4}: If $N_v \le |S|$, output the mesh and stop; otherwise, go to Step 2 with the updated $V_p$ and $V$.  

In our computations, we choose metric tensor $\MM_{aniso,3}$ in the above procedures for GPRAMA due to its good representation quality as described in Section \ref{Sec-AMArep}. In each iteration of Step 2, except the first, we only need to compute the significance measures for the neighboring vertices of $v_i$ after it is removed. Other vertices in the mesh outside of $\overline{\omega_i}$ are not affected. For efficient implementation, we do not need to delete the information at vertex $v_i$ such as coordinates, function value and neighboring triangles from the mesh data then triangulate the patch using the updated mesh data. We just need to replace the old triangulation of the patch containing $v_i$ with the new one without $v_i$ by updating the corresponding entries in the mesh data directly. A sorted index array for the significance measures can be used in Step 3 for efficient selection of the vertex to be removed.  

\subsection{Results}
\label{Subsec-result}

\begin{figure}[ht!]
\centering
\includegraphics[width=6in]{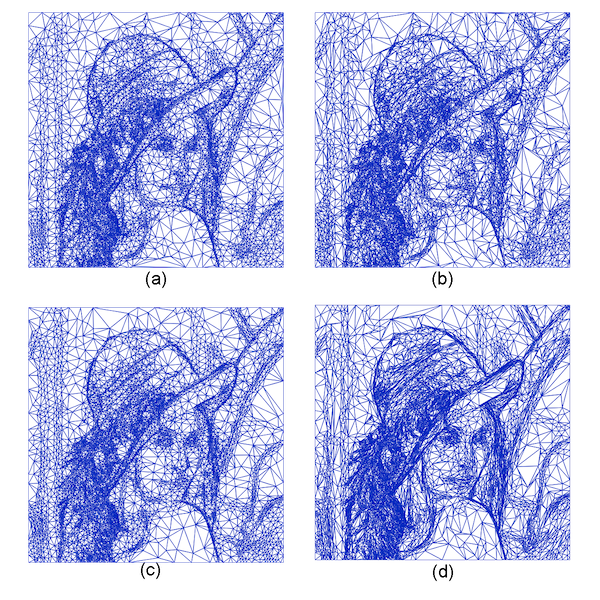}
\caption{Meshes of different GPR-related representations for image Lena at sample density of 3\%: (a) GPRED(4)-CDT, PSNR=33.49; (b) GPRED(4)-EC, PSNR=33.85; (c) GPRAMA(4)-CDT, PSNR=33.18; (d) GPRAMA(4)-EC, PSNR=34.51.}
\label{img-Lena-GPR}
\end{figure}

\begin{figure}[ht!]
\centering
\includegraphics[width=6in]{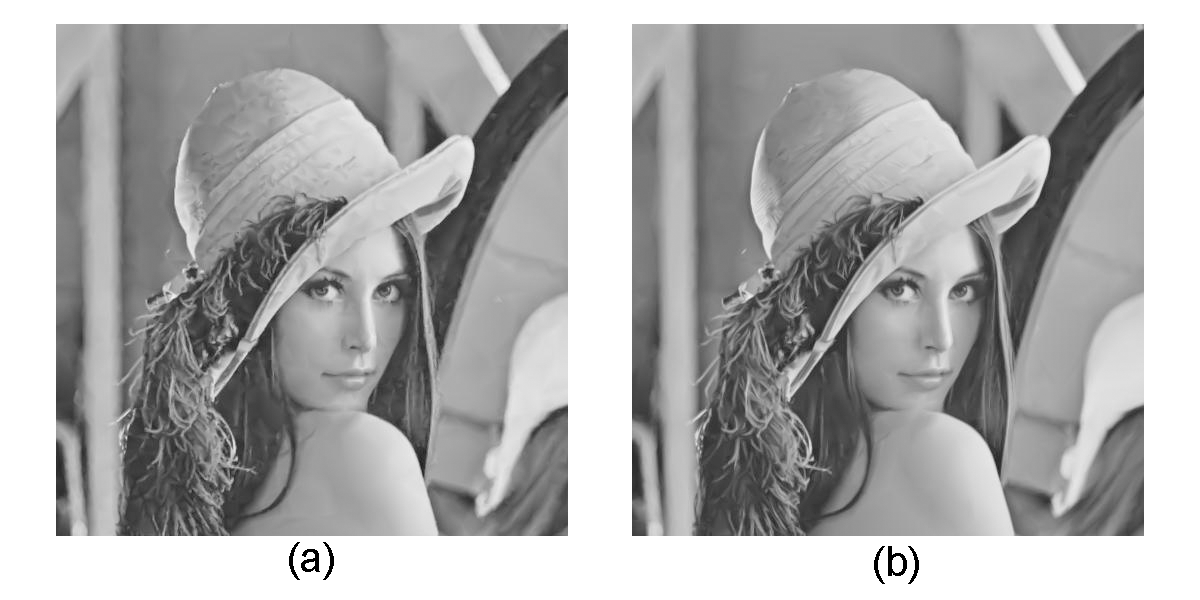}
\caption{Reconstructed images using different GPR-related representations for image Lena at sample density of 3\%: (a) GPRED(4)-EC, PSNR=33.85; (b) GPRAMA(4)-EC, PSNR=34.51.}
\label{img-Lena-GPR-2}
\end{figure}

According to the different triangulation methods for the mesh patching technique in Step 3, the final mesh and the corresponding representation are denoted as GPRAMA($\gamma$)-CDT if constrained Delaunay triangulation is used and GPRAMA($\gamma$)-EC if Ear Clipping method is used, where $\gamma$ specifies the number of initial points $|S_0|=\gamma |S|$. Figure \ref{img-Lena-GPR} shows the meshes obtained using GPRED and GPRAMA methods for image Lena at sample density 3\% with $\gamma=4$, and Figure \ref{img-Lena-GPR-2} presents two of the reconstructed images. 

\begin{figure}[ht!]
\centering
\includegraphics[width=6in]{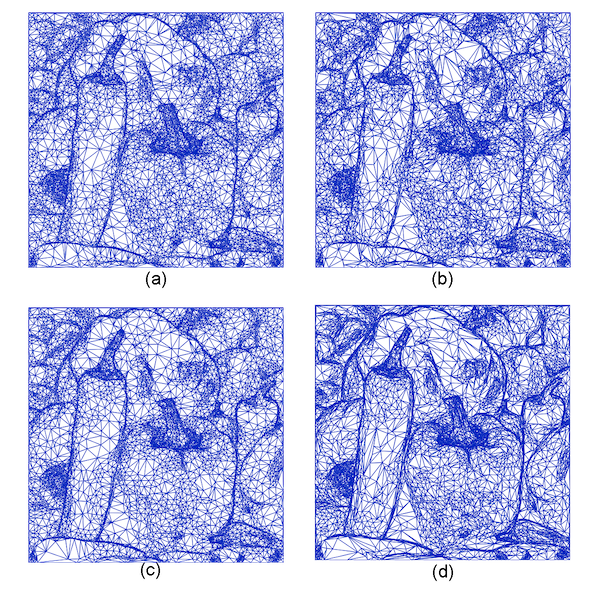}
\caption{Meshes of different GPR-related representations for image peppers at sample density of 3\%: (a) GPRED(4)-CDT, PSNR=33.65; (b) GPRED(4)-EC, PSNR=33.74; (c) GPRAMA(4)-CDT, PSNR=33.44; (d) GPRAMA(4)-EC, PSNR=34.23.}
\label{img-peppers-GPR}
\end{figure}

\begin{figure}[ht!]
\centering
\includegraphics[width=6in]{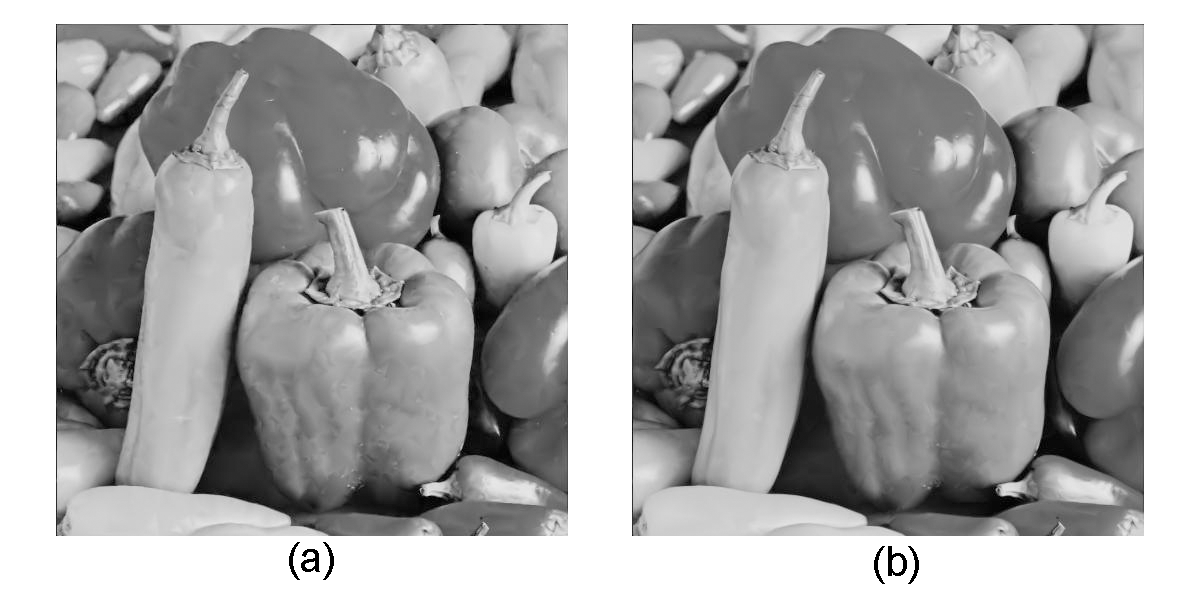}
\caption{Reconstructed images using different GPR-related representations for image peppers at sample density of 3\%: (a) GPRED(4)-CDT, PSNR=33.65; (b) GPRAMA(4)-EC, PSNR=34.23.}
\label{img-peppers-GPR-2}
\end{figure}

GPRED(4)-CDT has quality PSNR=33.49 while GPRED(4)-EC has PSNR=33.85. GPRAMA(4)-CDT has quality PSNR=33.18 while GPRAMA(4)-EC has PSNR=34.51. The GPRED-CDT method is essentially the GPRFS-ED method in Adam's paper \cite{Ada11} while GPRED-EC is an improved version of GPRED-CDT due to the different triangulation of the local patch. For GPRAMA, using constrained Delaunay triangulation for mesh patching does not preserve the anisotropy of the initial mesh, especially when significant amount of points are removed. Therefore, Ear Clipping method works better for GPRAMA, and GPRAMA-EC gives the best quality among the four GPR-related representations. Similar results are observed for image peppers as shown in Figures \ref{img-peppers-GPR} and \ref{img-peppers-GPR-2}. 

The mesh qualities at different sample densities for both Lena and peppers using GPR-related representations are shown in Table \ref{table-GPR}, where mesh patching with Ear Clipping are used for GPR and GPRAMA methods. GPRAMA(4) provides better quality than the traditional GPR method for both images except at sample density of $1\%$ for image peppers. Furthermore, GPRAMA(3) performs better than GPRED(5)-CDT and is comparable with GPRED(5)-EC for both images. Therefore, GPRAMA can achieve better quality than GPRED while starting with smaller $|S_0|$ which indicates less computational cost, especially for high-resolution images. The computational costs for different methods are provided in the next subsection. 

\begin{table}[thb]
\caption{Comparison of mesh qualities obtained with various GPR-related methods}
\vspace{2pt}
\centering
\begin{tabular}{|c|p{12mm}||p{10mm}|p{16mm}|p{16mm}|p{12mm}|p{12mm}|p{12mm}|}
\hline
& Sample & \multicolumn{6}{c|}{PSNR (dB)} \\ \cline{3-8} 
Image & Density (\%) & GPR & GPRED(5) -CDT & GPRED(5) -EC & GPR- AMA(2) & GPR- AMA(3) & GPR- AMA(4) \\
\hline
 & 1.0 & 29.15 & 29.33 & 30.19 & 29.48 & 30.48 & {\bf 30.84} \\ 
 & 2.0 & 31.81& 32.03 & 32.69 & 32.13 & 32.90 & {\bf 33.23} \\ 
Lena & 3.0 & 33.35 & 33.52 & 34.12 & 33.50 & 34.21 & {\bf 34.51} \\ 
 & 4.0 & 34.43 & 34.60 & 35.14 & 34.45 & 35.11 & {\bf 35.39} \\ 
\hline
 & 1.0 & {\bf 31.12} & 30.22 & 30.65 & 29.31 & 30.53 & 31.06 \\ 
 & 2.0 & 33.10 & 32.61 & 32.86 & 32.18 & 32.86 & {\bf 33.24} \\ 
peppers & 3.0 & 34.01 & 33.75 & 33.96 & 33.41 & 33.98 & {\bf 34.23} \\ 
 & 4.0 & 34.61 & 34.47 & 34.73 & 34.08 & 34.61 & {\bf 34.88} \\ 
\hline 
\end{tabular} \\
\vspace{2pt}
\label{table-GPR}
\end{table}

\begin{figure}[ht!]
\centering
\includegraphics[width=6in]{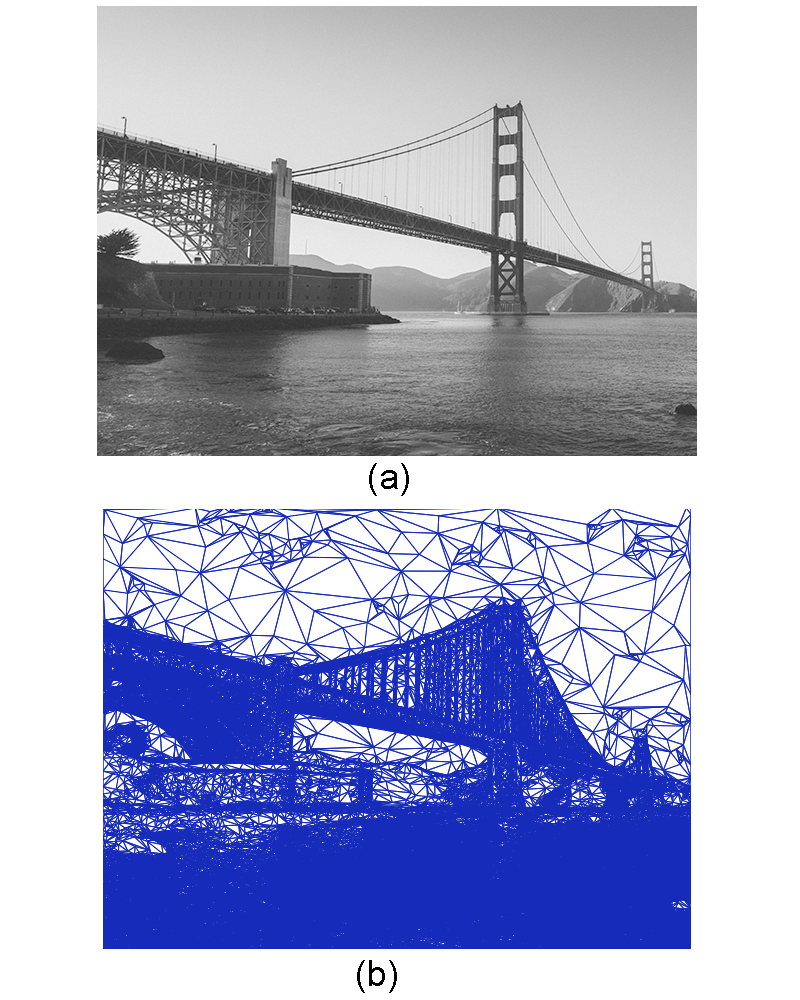}
\caption{GPRAMA representation for a image of Golden Gate bridge at sample density of 1\%: (a) original image, $4000 \times 3000$; (b) GPRAMA(4), PSNR=34.30.}
\label{img-bridge}
\end{figure}

\begin{figure}[ht!]
\centering
\includegraphics[width=6in]{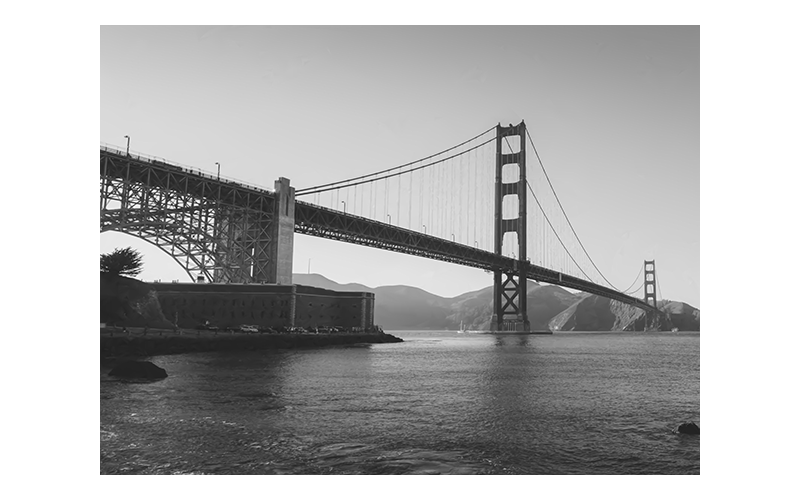}
\caption{Reconstructed image of GPRAMA(4) representation for Golden Gate bridge in Figure \ref{img-bridge}(a) at sample density of 1\%, PSNR=34.30.}
\label{img-bridge-2}
\end{figure}

Figure \ref{img-bridge} shows a picture of the Golden Gate bridge of pixel resolution $4000 \times 3000$ and its GPRAMA representation at sample density 1\% with $\gamma=4$. The reconstructed image is shown in Figure \ref{img-bridge-2} and the representation quality is PSNR=34.30. Some results using different representation methods are provided in Table \ref{table-bridge}, where Ear Clipping is used in the mesh patching for all the GPR-related methods. As can be seen, GPRAMA(2) has quality as good as GPRED(4) but starts with only half of the initial points. The results are consistent with our previous observations for other images.

\begin{table}[thb]
\caption{Comparison of mesh qualities for an image of Golden Gate bridge$^\dagger$}
\vspace{2pt}
\centering
\begin{tabular}{|p{20mm}||c|c|c|c|c|c|}
\hline
Sample & \multicolumn{5}{c|}{PSNR (dB)} \\ \cline{2-6} 
Density (\%) & ED & $\MM_{aniso,3}$ & GPRED(4) & GPRAMA(2) & GPRAMA(4) \\
\hline
 0.5 & 23.84 & 29.13 & 31.39 & 31.67 & {\bf 32.56} \\ 
 1.0 & 27.22 & 31.03 & 33.66 & 33.61 & {\bf 34.30} \\ 
\hline
\end{tabular} \\
\vspace{2pt}
$^\dagger$The image pixel resolution is $4000 \times 3000$.
\label{table-bridge}
\end{table}

\subsection{Computational complexity}
\label{Subsec-complexity}

Here, we compare the computational complexity of the various image representation methods considered in this paper. The computational complexity is
measured in terms of CPU execution time in seconds (converted from clock ticks) and varies for different hardware and software environment. Our computations in this paper are performed in a MacBook Pro laptop with 2.6GHz Intel Core i7 CPU, 8GB 1600MHz DDR3 memory, and OSX 10.9.5 operating system. 

Note that our program code was developed with basic level of efficiency and has not been optimized for execution speed. Thus the absolute CPU time for each method may be reduced by using highly optimized code. However, our focus here is to compare the computational complexity among the different methods. The CPU times for representations of images Lena and peppers with and without GPR are provided in Table \ref{table-cpu}, where Ear Clipping mesh patching technique is used in all the GPR-related methods. 

\begin{table}[thb]
\caption{Comparison of CPU time (in s) for different representations}
\vspace{2pt}
\centering
\begin{tabular}{|c|p{12mm}||p{10mm}|p{10mm}|p{12mm}||p{10mm}|p{16mm}|p{12mm}|}
\hline
& Sample & \multicolumn{3}{c||}{without GPR} & \multicolumn{3}{c|}{GPR-related} \\ \cline{3-8}
Image & Density (\%) & ED & $\MM_{H,3}$ & $\MM_{aniso,3}$ & GPR & GPRED(5) & GPR- AMA(3) \\
\hline
& 1.0 & 0.022 & 0.43 & 0.40 & 251 & 5.23 & 3.28 \\ 
& 2.0 & 0.024 & 0.42 & 0.44 & 249 & 6.97 & 3.97 \\ 
& 3.0 & 0.026 & 0.45 & 0.47 & 248 & 10.5 & 4.93 \\ 
Lena & 4.0 & 0.030 & 0.49 & 0.51 & 247 & 14.2 & 6.54 \\ 
& 6.0 & 0.033 & 0.54 & 0.57 & 245 & 28.9 & 11.06 \\ 
& 9.0 & 0.042 & 0.65 & 0.69 & 243 & 54.6 & 20.52 \\ 
& 12.0 & 0.049 & 0.76 & 0.80 & 240 & 82.8 & 37.36 \\ 
\hline
& 1.0 & 0.023 & 0.38 & 0.40 & 321 & 4.89 & 2.97 \\ 
& 2.0 & 0.024 & 0.40 & 0.42 & 319 & 6.73 & 3.70 \\ 
& 3.0 & 0.027 & 0.44 & 0.47 & 318 & 9.92 & 4.54 \\ 
peppers & 4.0 & 0.029 & 0.47 & 0.51 & 317 & 14.0 & 6.00 \\ 
& 6.0 & 0.039 & 0.53 & 0.60 & 315 & 25.7 & 10.6 \\ 
& 9.0 & 0.043 & 0.63 & 0.68 & 313 & 53.6 & 15.4 \\ 
& 12.0 & 0.050 & 0.72 & 0.76 & 309 & 93.0 & 34.7 \\ 
\hline 
\end{tabular} \\
\vspace{2pt}
\label{table-cpu}
\end{table}

It is clear that ED method is the fastest (with low quality) and GPR is the most computationally expensive method (with good quality), and the result is consistent with the existing literatures. For example, at sample density of 3\% for image Lena, ED representation takes only 0.026s, while GPR method takes 248s. For image peppers at sample density of 3\%, ED method takes 0.027s while GPR takes 318s. The smaller sample density, the longer time GPR needs because more points need to be removed before reaching the desired sample density. For all other methods, the computational cost is lower for smaller sample density. 

For representation methods without greedy-point removal technique, $\MM_{H,3}$ and $\MM_{aniso,3}$ provide better quality (see Table \ref{table-AMA}) but have higher computational cost than ED method (see Table \ref{table-cpu}). For example, at sample density of 3\% for image Lena, the $\MM_{H,3}$ representation takes 0.45 seconds and $\MM_{aniso,3}$ takes 0.47 seconds, which are about 16 times more than ED method but only 0.2\% of the time needed by GPR method. Similar results are observed for image peppers. Therefore, $\MM_{H,3}$ and $\MM_{aniso,3}$ are good balances between ED and GPR. In the meantime, $\MM_{aniso,3}$ takes about 0.02 seconds longer than $\MM_{H,3}$ which is due to the extra time needed to compute $\MM_{aniso}$ \eqref{M-aniso} in addition to $\MM_{H}$ \eqref{H-abs}. However, the extra cost is negligible (less than 5\%). 

For GPR-related representation methods, both GPRED(5) and GPRAMA(3) provides comparable quality with GPR (see Table \ref{table-GPR}) but take much less time (see Table \ref{table-cpu}). For example, at sample density of 3\% for image Lena, GPRED(5) takes 10.5s and GPRAMA(3) takes 4.93s. GPRAMA(3) only takes 2\% of the time needed by GPR and less than half of the time needed by GPRED(5). Therefore, GPRAMA method can provide comparable quality with GPRED but with lower computational cost, which makes it another good balance between ED and GPR. The results for image peppers are also similar.


\section{Conclusions and comments}
\label{Sec-con}

Adaptive sampling has become popular in image representation, among which triangular meshes have gained much interest. One common approach is to develop schemes to choose proper sample points then connect the points to form a triangular mesh. Another approach is to generate and adapt the mesh directly to represent the image. In this paper, we have introduced a framework of anisotropic mesh adaptation methods to image representation. The AMA methods take the $\MM$-uniform mesh approach and use a metric tensor $\MM$ to control the triangular mesh. Firstly, an initial Delaunay triangular mesh is generated based on the desired sample density. Then the mesh is adapted using the software BAMG according to the provided metric tensor. Lastly, finite element interpolation is used to reconstruct the image from the mesh. The anisotropic metric tensor $\MM_{aniso}$ in \eqref{M-aniso} provides the best representation in this framework among the considered metric tensors. Note that the method proposed by Courchesne {\it et al.} \cite{CGDC07} (with minor modification) is a special case within this framework. 

Within the AMA representation framework, we have developed a GPRAMA method based on the greedy-point removal scheme and a mesh patching technique. The local polygon (may be concave) surrounding a mesh vertex can be triangulated using constrained Delaunay triangulation or Ear Clipping method. When choosing the initial points using the error-diffusion scheme and CDT is chosen for mesh patching, the corresponding representation method is denoted as GPRED-CDT that is essentially the same as the GPRFS-ED method proposed by Adams \cite{Ada11}; while choosing EC for mesh patching leads to an improved version denoted as GPRED-EC. When starting the initial points from an AMA mesh, in particular, an $\MM_{aniso}$ mesh, and EC is chosen for mesh patching, we obtain the GPRAMA representation method that provides better quality than the GPRFS-ED method but with lower computational cost. Overall, mesh patching with EC provides better quality for GPR-related representation methods than mesh patching with CDT. Numerical results on two standard test images, Lena and peppers are presented, as well as on an image of the Golden Gate bridge that has higher resolution. The observations are confirmed by the results from three other images. All the results demonstrate that AMA representation is superior than ED representation, and GPRAMA performs better than the GPRFS-ED method.          

AMA representation methods have clear mathematic framework and provides flexibility for both adaptation using different metric tensors and reconstruction using different interpolation methods, although we have only focused on linear finite element interpolation in this paper. The AMA representation of the image and the mesh adaptation strategy will be useful for image scaling and PDE-based image processing such as image smoothing and edge enhancement using anisotropic diffusion filters, which are topics under our current investigation.

\section*{Appendix: Finite Element Interpolation for Triangles}
\label{Sec-app}

This appendix provides a brief introduction to the linear and quadratic finite element interpolation for triangles. The interpolation for quadrilateral elements are similar.  Fig. \ref{feminter} shows a triangular element $K$ and an isosceles right triangle $\hat{K}$ as the reference element. The vertices of $K$ are denoted as ${\bf a}_1$, ${\bf a}_2$ and ${\bf a}_3$, and the midpoints of the corresponding sides are denoted by ${\bf a}_4$, ${\bf a}_5$ and ${\bf a}_6$. The vertices of the reference element $\hat{K}$ are located at $\hat{{\bf a}}_1 (0,0)$, $\hat{{\bf a}}_2 (1,0)$, and $\hat{{\bf a}}_3 (0,1)$. The midpoints in $\hat{K}$ are located at $\hat{{\bf a}}_4 (0.5,0)$, $\hat{{\bf a}}_5 (0.5,0.5)$, and $\hat{{\bf a}}_6 (0,0.5)$ .

Denote the coordinates of the vertices of $K$ as ${\bf a}_1 (x_1,y_1)$, ${\bf a}_2 (x_2,y_2)$, and ${\bf a}_3 (x_3,y_3)$. The corresponding function values are denoted by $f_1 = f(x_1,y_1)$, $f_2 = f(x_2,y_2)$ and $f_3 = f(x_3,y_3)$. For any point ${\bf a}(x,y)$ in the element $K$, the corresponding point $\hat{{\bf a}}(\xi,\eta)$ in the reference element $\hat{K}$ is given by
\beq
 \left[ \begin{array}{c} \xi \\ \eta \end{array} \right] = \begin{bmatrix} x_2-x_1 & x_3-x_1 \\ y_2-y_1 & y_3-y_1 \end{bmatrix}^{-1} \times \left[ \begin{array}{c} x-x_1 \\ y-y_1 \end{array} \right].
\eeq

For linear interpolation, only the values at the three vertices are needed, and the function value at any point $(\xi,\eta)$ is interpolated as follows
\beq
f(\xi,\eta) = \sum_{i=1}^3 f_i \cdot N_i(\xi,\eta),
\label{int-lin}
\eeq
where $N_i(\xi, \eta)$ is the basis functions at $\hat{{\bf a}}_i$ and is defined as follows
\beq
N_1(\xi,\eta) = 1- \xi -\eta; \quad N_2(\xi,\eta)=\xi; \quad N_3(\xi,\eta)=\eta.
\label{basis-lin}
\eeq

For quadratic interpolation, the midpoints of the sides are needed. Denote the function values at midpoints ${\bf a}_4$, ${\bf a}_5$, and ${\bf a}_6$ as $f_4$, $f_5$, and $f_6$, respectively. Then the function value at any point $(\xi,\eta)$ in $\hat{K}$ is interpolated as follows
\beq
f(\xi,\eta) = \sum_{i=1}^6 f_i \cdot N_i(\xi,\eta),
\label{int-quad}
\eeq
where the basis functions are defined as follows
\bey
\nn
&& N_1 = (1-\xi-\eta)(1-2\xi-2\eta); \quad N_2 = \xi (2\xi-1); \\
\nn
&& N_3 = \eta (2\eta-1); \quad N_4 = 4 \xi (1-\xi-\eta); \\
&& N_5 = 4 \xi \eta; \quad N_6 = 4 \eta (1-\xi-\eta).
\label{basis-quad}
\eey

\vspace{14pt}
\noindent {\bf \Large Acknowledgement} \\
This work was partially supported by the grant from the University of Missouri Research Board (UMRB).

\end{document}